\documentclass[runningheads]{llncs}

\usepackage{eccv}

\usepackage{eccvabbrv}

\usepackage{graphicx}
\usepackage{booktabs}
\usepackage{xcolor}
\usepackage{lipsum}

\usepackage{bbm}

\usepackage[ruled,vlined]{algorithm2e}
\definecolor{commentcolor}{RGB}{110,154,155}   
\definecolor{asparagus}{rgb}{0.53, 0.66, 0.42}
\newcommand{\PyComment}[1]{\ttfamily\textcolor{commentcolor}{\# #1}}  
\newcommand{\PyCode}[1]{\ttfamily\textcolor{black}{#1}} 

\usepackage[accsupp]{axessibility}  

\usepackage[pagebackref,breaklinks,colorlinks,citecolor=eccvblue]{hyperref}

\usepackage{orcidlink}
\usepackage{color}

\definecolor{darkgreen}{RGB}{0,128,0}   

\usepackage{CJKutf8}

\usepackage{pifont}
\newcommand{\xmark}{\ding{55}}%

\usepackage{threeparttablex}
\usepackage{tabularx}
\newcolumntype{Y}{>{\centering\arraybackslash}X}

\begin{document}

\title{Multi Positive Contrastive Learning with Pose-Consistent Generated Images}

\titlerunning{GenPoCCL}

\author{Sho Inayoshi\inst{1} \and
Aji Resindra Widya\inst{1} \and
Satoshi Ozaki\inst{1} \and \\
Junji Otsuka\inst{1} \and
Takeshi Ohashi\inst{1}}

\authorrunning{S.Inayoshi et al.}

\institute{Sony Group Corporation, Tokyo, Japan \\
\email{\{Sho.Inayoshi, Aji.Widya, Satoshi.Ozaki, Junji.Otsuka, Takeshi.A.Ohashi\}@sony.com}}

\maketitle

\begin{abstract}
  Model pre-training has become essential in various recognition tasks.
  Meanwhile, with the remarkable advancements in image generation models, pre-training methods utilizing generated images have also emerged given their ability to produce unlimited training data.
  However, while existing methods utilizing generated images excel in classification, they fall short in more practical tasks, such as human pose estimation.
  In this paper, we have experimentally demonstrated it and propose the generation of visually distinct images with identical human poses. 
  We then propose a novel multi-positive contrastive learning, which optimally utilize the previously generated images to learn structural features of the human body. 
  We term the entire learning pipeline as GenPoCCL.
  Despite using only less than 1\% amount of data compared to current state-of-the-art method, GenPoCCL captures structural features of the human body more effectively, surpassing existing methods in a variety of human-centric perception tasks.  
\end{abstract}

\begin{figure}[ht]
  \begin{center}
    \includegraphics[width=\textwidth]{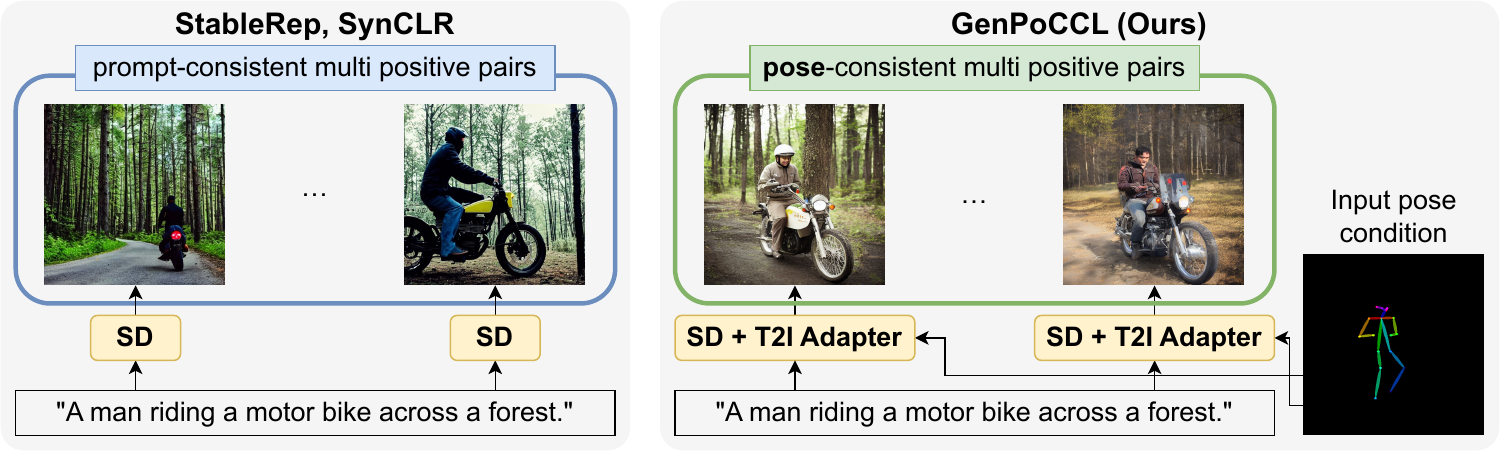}
    \caption{
      We compare our method to StableRep\cite{tian2023stablerep} and SynCLR\cite{tian2023synclr}.
      Both StableRep and SynCLR use a single prompt to generate semantically similar images which are then treated as positive pairs for contrastive learning.
      On the other hand, our GenPoCCL takes a step further by generating similar images from a same prompt and \textbf{same human body pose condition} for positive pairs for contrastive learning.}
    \label{fig:overview}
  \end{center}
\end{figure}

\section{Introduction} \label{sec:intro}
  Model pre-training has become essential in various recognition tasks.  
  Especially within human-centric perception tasks (\eg, human pose estimation, person ReID, \etc), there is a movement towards leveraging a pre-trained model as a foundational framework.
  This approach is motivated by the extensive variety of tasks, each with substantial learning costs, and the demonstrated superior performance of such models\cite{chen2023beyond,chen2022liftedcl,ci2023unihcp,hong2022versatile,tang2023humanbench,yuan2023hap}.
  Recently, HAP\cite{yuan2023hap} proposed to implicitly learn structural features of the human body, which has shown promising results.
  While pre-training shows promising results, it needs a huge amount of data.
  Hence, an increasing number of pre-training methods utilizing generated data have been proposed given their ability to produce unlimited training data\cite{tian2023stablerep,tian2023synclr,hammoud2024synthclip}.

  Meanwhile, image generation technology has advanced swiftly in recent years, achieving a quality that humans can hardly differentiate from actual images\cite{dhariwal2021diffusionbeatsgan}.
  Alongside this trend, recent studies\cite{tian2023stablerep,tian2023synclr,hammoud2024synthclip} have explored the pre-training of using generated images, and it has been reported that they achieve comparable or even superior performance to models pre-trained with real images.
  Thus, while pre-training with generated images indicates potential benefits, its effectiveness for tasks other than image classification and linear probing, particularly in practical scenarios, has not been confirmed.
  In addition to the rapid advancement in the quality of generated images, its controllability has also improved.
  Some researches\cite{mou2023t2i,li2023gligen,zhang2023adding} demonstrate the ability to control generated images in various methods, including line drawings, depth images, and human pose labels.
  Recent pre-training methods using generated images exploit the ability to generate consistent images from prompts, but the potential for consistent image generation with conditions other than prompts remains unexplored.
  
  In this paper, we tackle the challenge of suboptimal performance in human-centric perception tasks, a common shortfall of pre-training methods using generated images, by proposing a novel method named GenPoCCL. 
  The acronym GenPoCCL represents \textbf{Gen}erated image leveraged \textbf{Po}se \textbf{C}onsistent \textbf{C}ontrastive \textbf{L}earning.
  In GenPoCCL, we take advantage of the ability to generate images with uniform poses yet diverse appearances.
  By treating images generated from the same human body pose condition as positive pairs as shown in \cref{fig:overview},
  we propose pose-consistent multi-positive contrastive learning 
  to guide the model with human body pose constraints.
  We additionally introduce a special token named [POSE] token which allows to learn both discriminative human features and human-pose related features.
  Through GenPoCCL, images of the same pose can be mapped to proximate locations in the feature space, acquiring robust feature representations independent of background or human appearance.
  Consequently, this enables easier learning across human-centric perception tasks.
  Remarkably, it achieves superior performance to current methods with under 1\% of the generative data previously needed.

  The contributions of this research are as follows:
  \begin{itemize}
    \item We propose generation of visually distinct images with identical human poses for self-supervised pre-training.
    \item We introduce GenPoCCL, a novel multi-positive contrastive learning approach that utilizes the generated pose-consistent appearance-varying images. 
    It allows the model to learn structural features of the human body.
    \item We also introduce [POSE] token in GenPoCCL, which allows to learn both discriminative human features and human-pose related features.
    \item We experimentally show that GenPoCCL surpasses existing methods in a variety of human-centric perception tasks, with using only less than 1\% amount of data compared to existing methods.
  \end{itemize}

\section{Related works} \label{sec:related}
\subsection{Image generation} \label{subsec:imggen}
Data has been always an important part of training a neural network. 
Depending on the task, a massive amount of data is needed in which obtaining them would become a tremendous task.
On the other hand, generative models such as Normalizing Flow\cite{rezende2015variational,kingma2018glow,dinh2016density,dinh2014nice}, Generative Adversarial Network (GAN)\cite{goodfellow2020generative,ledig2017photo,brock2018large}, and especially diffusion models\cite{rombach2022stablediffusion,ho2020ddpm,song2020ddim,song2019generative,song2020improved,song2020score}, 
have been showing a massive leap in generation quality, pushing learning with generated data to become the norm\cite{sariyildiz2023fakeittillyoumakeit,azizi2023synthetic,chen2019learning,pumarola20193dpeople,he2022synthetic}.
In recent years, diffusion models have gained popularity through their novel noise-denoising approach to generate highly detailed and realistic images.
In addition, it has also been proven that diffusion model consistently beats other generative models in terms of generation quality\cite{dhariwal2021diffusionbeatsgan}.
Considering the importance of image quality in our work, we decide to opt for the diffusion model.

Diffusion models also offer flexibility in generation as they can be conditioned on various modalities.
To generate a specific image, the most common way to guide the generation process is to use text as a guidance. Unfortunately, text-only guidance cannot capture the full preference on how the generated image should look like.
To overcome this challenge, recent studies grant enhanced controllability in generation process by injecting extra features as guidance extracted from an external control module\cite{mou2023t2i,zhang2023adding,li2023gligen}.
For example, T2I-adapter\cite{mou2023t2i} and ControlNet\cite{zhang2023adding} make it possible to generate image of human in specific pose by giving an OpenPose\cite{cao2017openpose} bone image as input condition.
Additionally, GLIGEN\cite{li2023gligen} allows the direct utilization of keypoint coordinates as input condition.
Greater precision in controlling the image generation means more complex and better synthetic data can be obtained.

In this paper, we take advantage of controllable image generation to generate synthetic data consisting of pose-consistent, appearance-varying individuals. 
We then show how to effectively use them to train a neural network for representation learning task.

\subsection{Representation learning} \label{subsec:replearn}
Representation learning is a learning methodology that allows a network to automatically discover meaningful representation based on the structure and relation of the provided training data.
It focuses on identifying patterns and features that are important for understanding the data in a self-supervised manner.
In recent years, masked image modeling (MIM) and contrastive learning (CL) emerged as the mainstream representation learning approach. 
MIM\cite{he2022mae,bao2021beit,peng2022beitv2,chen2024autoencodercontextlearning,he2020momentum,wei2022maskedfeatureprediction} is a generative approach that learns representations by reconstructing the high portion of corrupted part in the given input based on the remaining information.
On the other hand, CL\cite{caron2021dino,chen2020mocov2,grill2020byol,chen2021mocov3} is an approach which learns representations by discriminating between the similarity (dissimilarity) of positive (negative) image pairs via contrastive loss.
Hence, having good positive and negative image pairs is very important for CL. Given a single original image, applying strong augmentations such as crop, resize, and color jitter, has been the standard to create multiple positive pairs\cite{chen2020simclr}.
However, recent studies show that creating positive pairs from different-yet-semantically-close data leads to better representation learning\cite{radford2021clip,tian2023stablerep,el2023learning}.

Recently, StableRep\cite{tian2023stablerep} has shown that Stable Diffusion\cite{rombach2022stablediffusion} can be used to generate infinite variations of images to be treated as positive image pairs for CL.
SynCLR\cite{tian2023synclr} also proposes a pre-training method with a large-scale exclusively synthetic dataset constructed by generating images from LLM-created prompts.
Similar to StableRep, they generate multiple semantically similar but visually diverse images from a single prompt, serving as multi-positive pairs for CL.
SynthCLIP\cite{hammoud2024synthclip} also constructs a fully synthetic, large-scale text-image pairs dataset to train CLIP\cite{radford2021clip} model.
Unfortunately, even though they show promising results, the effectiveness beyond classification and linear probing is still unconfirmed. 
Broader application of utilizing synthetic data for representation learning that target real life cases is yet to be explored. 

Given its significance in numerous real-world scenarios, addressing human-centric understanding has emerged as a crucial challenge.
While ImageNet\cite{deng2009imagenet} pre-training has been the standard, recent studies show that self-supervised pre-training approaches using a collection of human images yields improved understanding of human-centric attributes by the network\cite{yuan2023hap,chen2022spot,hong2022hcmoco,ci2023unihcp,tang2023humanbench}.
Among these studies, HAP\cite{yuan2023hap} shows promising results by proposing to incorporate human-part prior to guide masking strategy during MIM-based representation learning.
In short, they select human body parts at random and mask the image patches that correspond to these selected areas.
In addition, CL-equivalent structure-invariant alignment loss is also proposed to improve the network ability to capture the characteristic of human body.

Motivated by HAP success, we are encouraged to further improve it by utilizing synthetic data for the pre-training task.
In our work, we exploit fine-grained controllable generation to generate specifically crafted images by fixing specific parts of the generated images, \eg, varying human appearance with fixed pose.
We then show how our proposed multi-positive CL is able to efficiently utilize the specifically crafted generated image pairs to enhance the network understanding of better representations.

\begin{figure}[t!]
  \begin{center}
    \includegraphics[width=\textwidth]{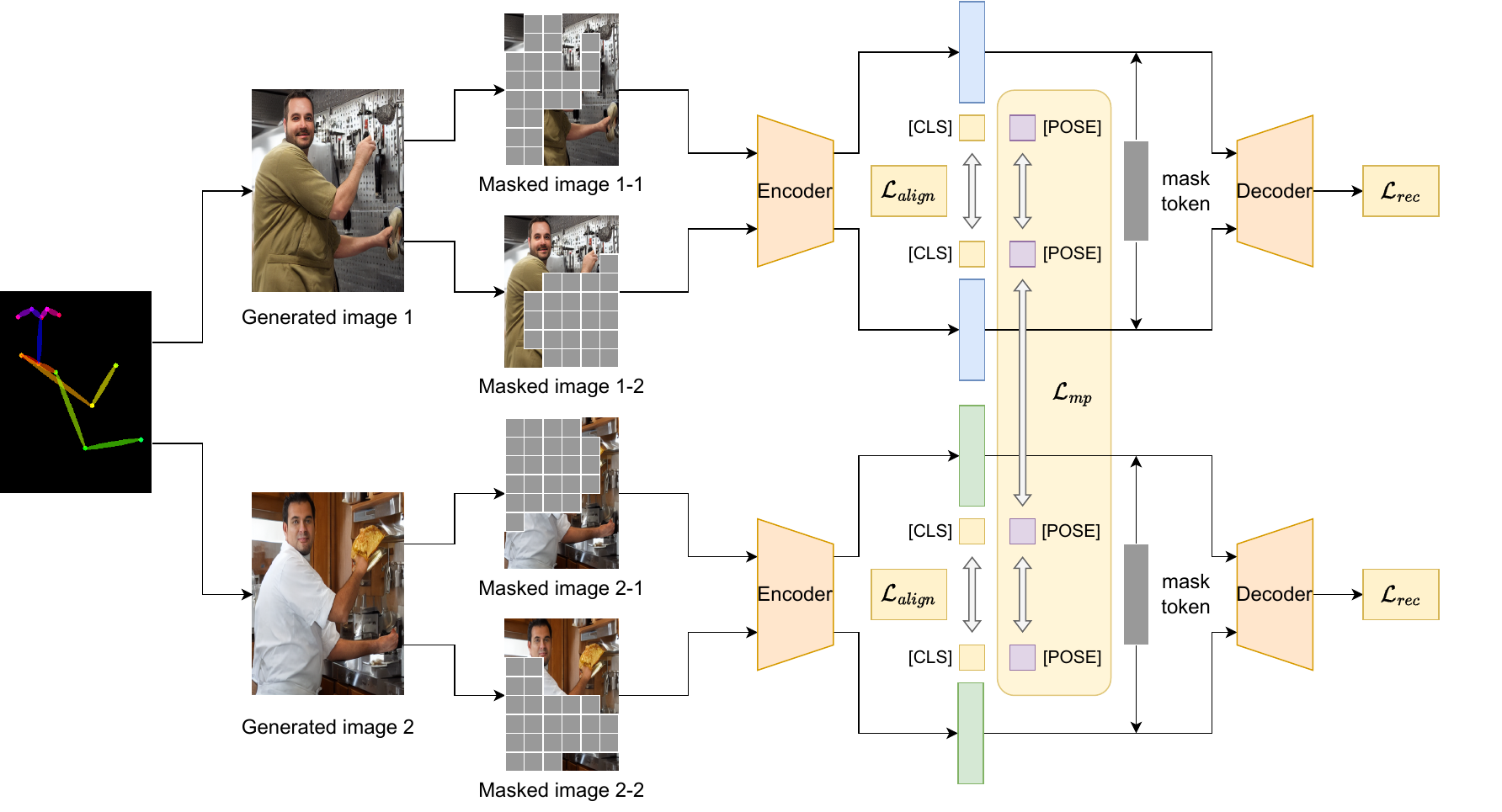}
    \caption{
      Overall pipeline of our method. 
      Utilizing Stable Diffusion\cite{rombach2022stablediffusion} and T2I-Adapter\cite{mou2023t2i}, 
      we generate pose-consistent images with varying appearances for contrastive learning from a single prompt and pose condition by altering the initial seed.
      In pre-training, selected human-part patches are masked following HAP\cite{yuan2023hap}. 
      We reconstruct images using a shared encoder-decoder, applying reconstruction loss $\mathcal{L}_{rec}$. 
      Alignment of [CLS] tokens is achieved with HAP's alignment loss $\mathcal{L}_{align}$, 
      and we introduce a [POSE] token with a multi-positive contrastive loss $\mathcal{L}_{mp}$ to refine pose and appearance learning.
    }
    \label{fig:pipeline}
  \end{center}
\end{figure}

\section{Proposed method} \label{sec:propose}
In our research, we employed generative models to generate images consistent with human body pose conditions, which are then utilized in self-supervised pre-training.
Our training pipeline, detailed in \cref{fig:pipeline}, begins by generating multiple images from a single human pose label, as elaborated in \cref{subsec:t2i}.
Subsequent steps include applying different masks in conjunction with the human body pose labels and processing these through a weight-shared encoder to extract features.
We employ a [CLS] token for aligning features of the same identity using an alignment loss $\mathcal{L}_{align}$ proposed in HAP\cite{yuan2023hap},
and introduce a [POSE] token to align features of people with consistent poses using a multi-positive contrastive learning objective $\mathcal{L}_{mp}$, which is further described in \cref{subsec:pcmpcl}.

A simple alternative approach is to utilize input conditions and corresponding generated images as direct paired training data for downstream tasks.
However, given the current performance of controllable generation models, generating a perfectly consistent image with the input condition remains a challenge,
making them inadequate for supervised learning in downstream tasks that demand exact labels.
On the other hand, as mentioned in \cref{subsec:replearn}, HAP shows that indirect utilization of human body pose labels leads to better understanding of structural features of the human body.
For this application, the current performance of controllable generation models is sufficient.
Therefore, we leverage generated data as training data for self-supervised pre-training.
Further details will be discussed in \cref{subsec:main_result}.

\begin{figure}[t]
  \begin{center}
    \includegraphics[width=\textwidth]{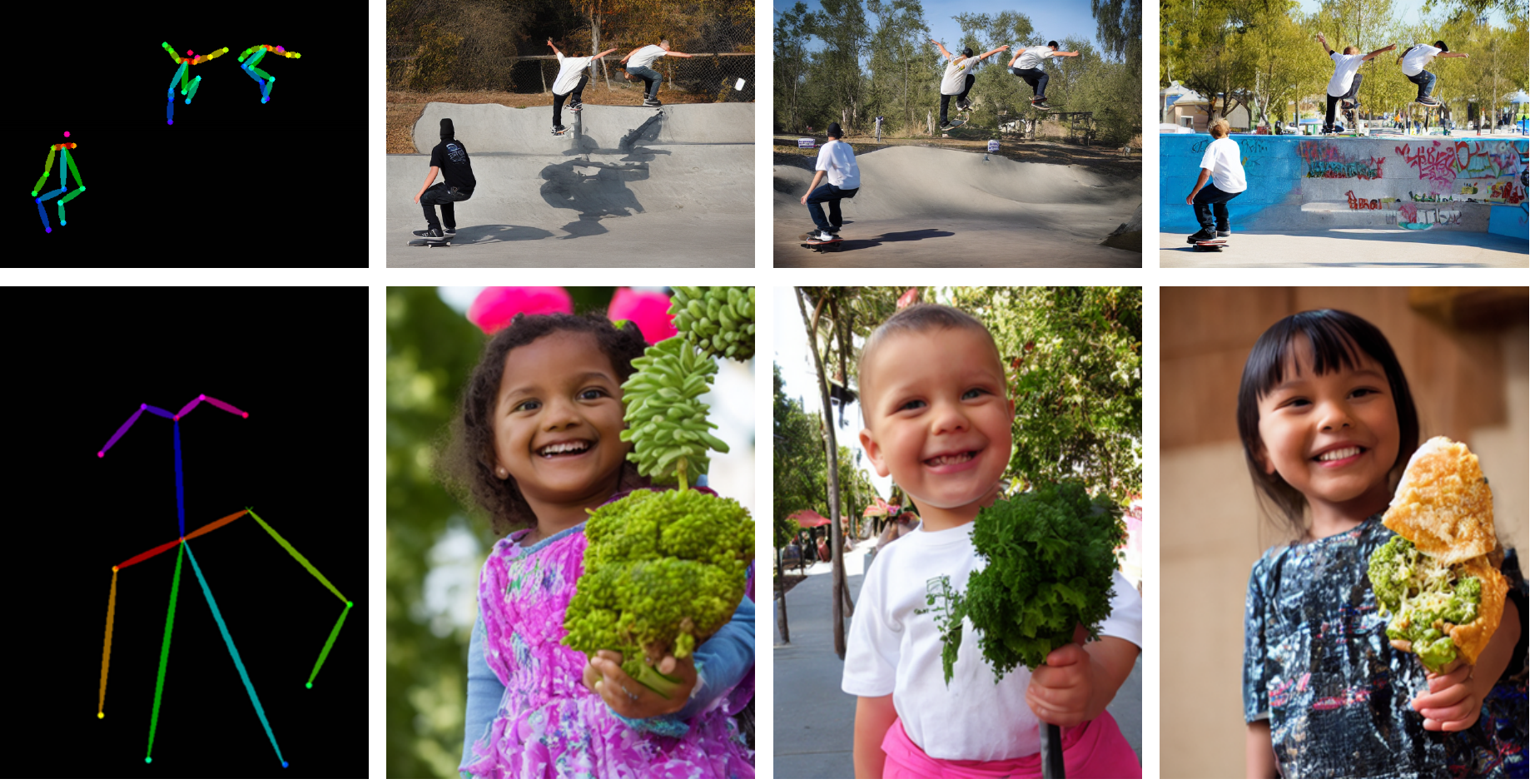}
    \caption{
      Examples of generated images with input human body pose condition.
      By varying the initial noise, we succeed in generating pose-consistent, appearance-varied images.
      We crop the images along with bounding-box labels.
    }
    \label{fig:example}
  \end{center}
\end{figure}

\subsection{Pose-conditioned image generation} \label{subsec:t2i}
  Motivated by the recent advancement in controllable image generation, we utilized the T2I-Adapter\cite{mou2023t2i} for the image generation model.
  T2I-Adapter adds a small network to a pre-trained diffusion model, enabling image generation under various conditions.
  Leveraging a publicly available controllable model with human body poses, we generated five distinct images for a single human pose since MSCOCO dataset\cite{lin2014microsoft} has five slightly different caption labels for each image.
  Inputs for image generation were human body pose and caption labels derived from the MSCOCO dataset's training split.
  Despite identical human body pose inputs, the use of varied initial noise generates images with appearance variations for each pose as shown in \cref{fig:example}. 
  The generated images are then cropped based on MSCOCO's bounding box label to extract human area.
  Notably, MSCOCO's labels for small human poses presented challenges, as the T2I-Adapter struggled to generate accurate representations.
  Consequently, bounding boxes smaller than 4,096 pixels were excluded.
  Furthermore, some MSCOCO labels correspond to incomplete human figures, which also prevented accurate human generation by the T2I-Adapter.
  Thus, images displaying fewer than five of the seventeen whole body keypoints were likewise omitted.
  The refined dataset consists of 117,126 unique human poses with five variations each were utilized for pre-training.
  In this paper, we refer to this generated dataset as GenCOCO dataset.
  
  To investigate the impact of increased data scale, we also created a large-scale generated dataset based on the LUPerson\cite{fu2020luperson} dataset.
  We resized the human body pose pseudo-labels of images from the LUPerson dataset to 192 pixels in width, serving as input for image generation.
  Due to the absence of caption labels in the LUPerson dataset, a rule-based method was employed for caption generation, which was then used as input for image generation.
  This rule-based method is detailed in the supplementary material.
  As the data predominantly consists of full-body images, we do not perform any filtering.
  Consequently, 4,180,243 unique human poses with three variations each were utilized for pre-training.
  In this paper, we refer to this generated dataset as GenLUPerson dataset.

\subsection{Pose-consistent multi-positive contrastive learning} \label{subsec:pcmpcl}
  HAP\cite{yuan2023hap} suggests that implicitly learning structural features of the human body improves performance in human-centric perception tasks.
  Our proposed GenPoCCL applies multi-positive contrastive learning to bring features of images from the same pose closer together as shown in \cref{fig:pipeline}, 
  enhancing the understanding of structural human body features.
  This property is unique to controllable generative models that can generate visually distinct images with identical human poses, where obtaining such real image pairs is deemed to be impossible.
  By adopting this learning approach, we make the feature extractor learn to project images with identical poses into around the same area in the feature space,
  acquiring robust features that are not influenced by the background or appearance.
  Consequently, this enables easier learning across human-centric perception tasks.
  To better capture structural human body features, we enhance the HAP pipeline with two key elements: a [POSE] token and a multi-positive contrastive loss.

  \subsubsection{Pose token.} \label{subsec:posetk}
    HAP employs a structure-invariant alignment loss to align the [CLS] token across two variations of masked images.
    Inspired by this, we considered employing a [CLS] token for multi-positive contrastive learning.
    However, we found that letting the [CLS] token both learn discriminative human features and to seek feature alignment across different image sharing the same pose is not optimal. 
    Therefore, we proposed a new [POSE] token which tackles this problem.
    By incorporating the [POSE] token, we maintain the benefits of learning discriminative human characteristics via structure-invariant alignment loss 
    and further strengthen the acquisition of the structural human body features.
    In practice, we demonstrate improved learning progression and enhanced performance in human-centric perception tasks, as shown in \cref{subsec:ablation}.
    \begin{algorithm}[htb]
      \scriptsize
      \SetAlgoLined
        \PyComment{enc: encoder} \\
        \PyComment{dec: decoder} \\
        \PyComment{tau: temperature} \\
        \PyComment{minibatch x: [n, m, 3, h, w]} \\
        \PyComment{keypoints kpts: [n, m, 17, 2]} \\
        \PyComment{n poses, m images per pose} \\
        \PyCode{for x, kpts in loader:} \\
        \Indp   
          \PyComment{apply different appearance augmentation for each (n*m) images} \\
          \PyCode{x = appearance\_augment(x)} \\ 
          \PyComment{apply same geometric augmentation for same pose images} \\
          \PyCode{x = geometric\_augment(x)} \\ 
          \PyCode{x = cat(unbind(x, dim=1))}  \PyComment{[n*m, 3, h, w]} \\
          \PyComment{use keypoint labels to apply masks based on human body pose} \\
          \PyCode{mask1 = gen\_mask(cat(unbind(kpts, dim=1)))} \\
          \PyCode{mask2 = gen\_mask(cat(unbind(kpts, dim=1)))} \\
          \PyCode{x1 = x * mask1}  \PyComment{add mask to input image} \\
          \PyCode{h1 = enc(x1)} \\
          \PyCode{cls1 = h1[:, 0, :]}  \PyComment{extract [CLS] token} \\
          \PyCode{cls1 = normalize(cls1)} \\
          \PyCode{pose1 = h1[:, 1, :]}  \PyComment{extract [POSE] token} \\
          \PyCode{pose1 = normalize(pose1)} \\
          \PyCode{mask\_token1 = Parameter((n*m, d))}  \PyComment{d: embed dimension of decoder} \\
          \PyCode{rec1 = dec(h1, mask\_token1)} \\
          \PyCode{x2 = x * mask2} \PyComment{add another mask to input image} \\
          \PyCode{h2 = enc(x2)} \\
          \PyCode{cls2 = h2[:, 0, :]}  \PyComment{extract [CLS] token} \\
          \PyCode{cls2 = normalize(cls2)} \\
          \PyCode{pose2 = h2[:, 1, :]}  \PyComment{extract [POSE] token} \\
          \PyCode{pose2 = normalize(pose2)} \\
          \PyCode{mask\_token2 = Parameter((n*m, d))} \\
          \PyCode{rec2 = dec(h2, mask\_token2)} \\
          \PyComment{compute loss only for masked region} \\
          \PyCode{loss\_rec = 0.5 * (MSE(rec1, x, mask1) + MSE(rec2, x, mask2))} \\
          \PyCode{loss\_align = 0.5 * (InfoNCE(cls1, cls2) + InfoNCE(cls2, cls1))} \\
          \PyCode{loss\_mp = 0.5 * (MPCLoss(pose1, pose2) + MPCLoss(pose2, pose1))} \\
          \PyCode{loss = loss\_rec + $\gamma_1$ * loss\_align + $\gamma_2$ * loss\_mp} \\
          \PyCode{loss.backward()} \\
        \Indm 
      \caption{PyTorch-style pseudocode for our training pipeline}
      \label{algo:pipeline}  
    \end{algorithm}

  \subsubsection{Multi-positive contrastive loss.} \label{subsec:mpcl}
    Similar to StableRep\cite{tian2023stablerep}, we describe multi-positive contrastive learning as a matching problem.
    Suppose that there is an encoded anchor sample $\boldsymbol{a}$ and a collection of encoded candidates $\lbrace\boldsymbol{b}_1, \boldsymbol{b}_2, ..., \boldsymbol{b}_K\rbrace$.
    We calculate a contrastive categorical distribution $\mathbf{q}$ that indicates the likelihood of $\boldsymbol{a}$ matching each of $\boldsymbol{b}$:
    \begin{equation}
      \mathbf{q}_i = \frac{\exp(\boldsymbol{a} \cdot \boldsymbol{b}_i / \tau)}{\sum_{j=1}^{K}\exp(\boldsymbol{a} \cdot \boldsymbol{b}_j / \tau)}
    \end{equation}
    where $\tau$ represents a temperature hyper-parameter set to 0.2. Here, $\boldsymbol{a}$ and all $\boldsymbol{b}$ are $l_2$ normalized.
    Fundamentally, it can be viewed as a $K$-way softmax classification distribution over all encoded candidates.
    We can then interpret the ground-truth categorical distribution of $\mathbf{p}$ as:
    \begin{equation}
      \mathbf{p}_i = \frac{\mathbbm{1}_{\mathrm{match}(\boldsymbol{a}, \boldsymbol{b}_i)}}{\sum_{j=1}^{K}\mathbbm{1}_{\mathrm{match}(\boldsymbol{a}, \boldsymbol{b}_j)}}
    \end{equation}
    where the indicator function $\mathbbm{1}_{\mathrm{match}(\cdot,\cdot)}$ denotes whether the anchor and candidate match each other.
    Finally, the multi-positive contrastive loss can be calculated as the cross-entropy between the ground-truth distribution $\mathbf{p}$ and the contrastive distribution $\mathbf{q}$, which can be expressed as:
    \begin{equation}
      \mathcal{L}_{mp} = H(\mathbf{p}, \mathbf{q}) = -\sum_{i=1}^{K}\mathbf{p}_i\log\mathbf{q}_i
    \end{equation}

    This loss function is closely related to that used in StableRep, yet it differs by aligning features between images from the same human pose rather than the same prompt.
    The Pytorch-like pseudocode of the batched multi-positive contrastive learning algorithm is described in the supplementary material.

  \subsection{Overall training pipeline} \label{subsec:overall}
    The overall loss function can be expressed as:
    \begin{equation}
      \mathcal{L} = \mathcal{L}_{rec} + \gamma_1\mathcal{L}_{align} + \gamma_2\mathcal{L}_{mp}
    \end{equation}
    where $\mathcal{L}_{rec}$ is the MSE loss for image reconstruction following\cite{he2022mae}, 
    and $\mathcal{L}_{align}$ is the InfoNCE loss for aligning features under varying masks following\cite{yuan2023hap}.
    We use $\mathcal{L}_{align}$ and $\mathcal{L}_{mp}$ to align [CLS] and [POSE] tokens respectively.
    $\gamma_1$ and $\gamma_2$ are the weights to balance the three loss functions.
    We set $\gamma_1$ to 0.05 following\cite{yuan2023hap}, and also, $\gamma_2$ to 0.05 by default.
    The Pytorch-like pseudocode of our overall training pipeline is described in \cref{algo:pipeline}.
    Each batch consists of $n \times m$ images, meaning that we sample $m$ images for each of the $n$ poses.
    Here we apply data augmentation; appearance augmentations vary across all images, while geometric augmentations remain consistent for images derived from the same pose.
    This ensures that geometric augmentations do not alter the consistency of human poses across images from the same pose.

\section{Experiments}
\subsection{Settings} \label{subsec:settings}
\subsubsection{Pre-training.} 
  In our approach, as detailed in \cref{subsec:t2i}, the GenCOCO dataset and the GenLUPerson dataset are utilized for pre-training.
  Keypoint labels for both datasets are derived from the input human pose conditions used during image generation.
  The encoder model structure is based on the ViT-Base\cite{dosovitskiy2021vit}.
  For the GenCOCO dataset, we use an input resolution of 256 $\times$ 192  and a batch size of 2,560 (\ie, 64 poses $\times$ 5 variations $\times$ 8 GPUs).
  Similarly, for the GenLUPerson dataset, we maintain the same resolution and set the batch size to 6,144 (\ie, 256 poses $\times$ 3 variations $\times$ 8 GPUs).
  Due to GPU memory limitations, we could not load all mini-batches at once and thus resorted to gradient accumulation.
  We adopt AdamW\cite{loshchilov2018decoupled} as the optimizer in which the weight decay is set to 0.05.
  We use cosine decay learning rate scheduling\cite{loshchilov2017sgdr} with the base learning rate set to 2.4e-3.
  For experiments on the GenCOCO dataset, we set warmup epochs to 40 and total epochs to 800. 
  For the GenLUPerson dataset, we set warmup epochs to 1 and total epochs to 40. 
  The model parameters are initialized based on the Xavier initialization method\cite{pmlr-v9-glorot10a}.
  As mentioned in \cref{subsec:overall}, appearance augmentations vary across all images, while geometric augmentations remain consistent for images derived from the same pose.
  Compared to the common settings in prior research\cite{he2022mae,yuan2023hap}, we apply stronger augmentation to bridge the domain gap between real and generated images.
  We additionally adopt various types of augmentations such as random rotation, random blur, \etc.
  We empirically validate the effectiveness of stronger augmentation in \cref{subsec:ablation}.
  Details of specific augmentations are provided in the supplementary material.
  After pre-training, the encoder is retained along with task-specific heads to address the downstream human-centric perception tasks, while other modules are discarded.

\subsubsection{Human-centric perception tasks.}
  We perform quantitative evaluation on six benchmarks over four human-centric perception tasks. 
  These tasks include 2D human pose estimation on MPII\cite{andriluka2014mpii} and MSCOCO\cite{lin2014microsoft}, person ReID on Market-1501\cite{zheng2015market},
  text-to-image person ReID on RSTPReid\cite{zhu2021rstpreid}, and pedestrian attribute recognition on PA-100K\cite{liu2017pa100k} and PETA\cite{deng2014peta}. 
  For 2D pose estimation, we employ PCKh and AP as metrics for MPII and MSCOCO respectively. 
  Our framework is built upon the MMPose codebase\cite{mmpose2020}. 
  In person ReID, we adopt the method of Lu \etal\cite{lu2022improving} and report mAP. 
  For text-to-image person ReID, we report Rank-1 using the approach of Shao \etal\cite{shao2022learning}. 
  In pedestrian attribute recognition, we report mean accuracy (mA) with the strategy proposed by Jia \etal\cite{jia2021rethinking}. 
  Further training specifics are available in the supplementary material.

\begin{table}[htbp]
  \caption{\textbf{Main results.} We compare GenPoCCL with representative pre-training methods. \textit{Syn.} means utilization of synthetic data during pre-training. $^{\dag}$ indicates the results of our replication experiments using the pre-trained model and the implementation published by \cite{yuan2023hap}. We report mean performance over three independent trials.}
  \label{tab:results}
  \centering
  \begin{subtable}[t]{0.48\textwidth}
    \centering
    \subcaption{\textbf{Dataset information.} We compare the number of training data for each competing method.}
    \label{subtab:data_scale}
    \begin{tabular}{lcc}
      \toprule
      Method & Datasets & Samples \\
      \midrule
      HAP\cite{yuan2023hap} & LUPerson & 2.1M \\
      \midrule
      $\rm{StableRep_{CC}}$\cite{tian2023stablerep} & GenCC12M & 83M \\
      $\rm{StableRep_{Red}}$\cite{tian2023stablerep} & GenRedCaps & 105M \\
      SynCLR\cite{tian2023synclr} & SynCaps-150M & 600M \\
      SynthCLIP\cite{hammoud2024synthclip} & SynthCI-30M & 30M \\
      \midrule
      $\rm{GenPoCCL_{COCO}}$ & GenCOCO & 0.59M \\
      $\rm{GenPoCCL_{LU}}$ & GenLUPerson & 12M \\
      \bottomrule
    \end{tabular}
  \end{subtable}%
  \hfill
  \begin{subtable}[t]{0.48\textwidth}
    \centering
    \subcaption{\textbf{2D pose estimation.} PCKh($\%$) is reported for MPII and AP($\%$) is reported for MSCOCO.}
    \label{subtab:pose}
    \begin{tabular}{lccc}
      \toprule
      Method & Syn. & MSCOCO & MPII \\
      \midrule
      HAP\cite{yuan2023hap} & \xmark & 75.9 & 91.8 \\
      $\rm{HAP^{\dag}}$\cite{yuan2023hap} & \xmark & 75.8 & 91.2 \\
      \midrule
      $\rm{StableRep_{CC}}$\cite{tian2023stablerep} & \checkmark & 73.9 & 88.3 \\
      $\rm{StableRep_{Red}}$\cite{tian2023stablerep} & \checkmark & 74.0 & 88.0 \\
      SynCLR\cite{tian2023synclr} & \checkmark & \textbf{74.3} & 88.0 \\
      SynthCLIP\cite{hammoud2024synthclip} & \checkmark & 73.1 & 86.0 \\
      \midrule
      Supervised & \checkmark & 71.3 & - \\
      \midrule
      $\rm{GenPoCCL_{COCO}}$ & \checkmark & 74.1 & 89.2 \\
      $\rm{GenPoCCL_{LU}}$ & \checkmark & 74.2 & \textbf{89.7} \\
      \bottomrule
    \end{tabular}
  \end{subtable}%
  \hfill
  \begin{subtable}[t]{0.48\textwidth}
    \centering
    \subcaption{\textbf{Person re-id.} We evaluate on Market-1501 dataset\cite{zheng2015market}. mAP($\%$) is reported.}
    \label{subtab:reid}
    \begin{tabular}{lcc}
      \toprule
      Method & Syn. & Market-1501 \\
      \midrule
      HAP\cite{yuan2023hap} & \xmark & 91.7 \\
      $\rm{HAP^{\dag}}$\cite{yuan2023hap} & \xmark & 92.0 \\
      \midrule
      $\rm{StableRep_{CC}}$\cite{tian2023stablerep} & \checkmark & 70.5 \\
      $\rm{StableRep_{Red}}$\cite{tian2023stablerep} & \checkmark & 77.5 \\
      SynCLR\cite{tian2023synclr} & \checkmark & 86.6 \\
      SynthCLIP\cite{hammoud2024synthclip} & \checkmark & 81.5 \\
      \midrule
      $\rm{GenPoCCL_{COCO}}$ & \checkmark & 83.0 \\
      $\rm{GenPoCCL_{LU}}$ & \checkmark & \textbf{86.7} \\
      \bottomrule
    \end{tabular}
  \end{subtable}%
  \hfill
  \begin{subtable}[t]{0.48\textwidth}
    \centering
    \subcaption{\textbf{Text-to-image person re-id.} We evaluate on RSTPReid dataset\cite{zhu2021rstpreid}. Rank-1($\%$) is reported.}
    \label{subtab:t2i_reid}
    \begin{tabular}{lcc}
      \toprule
      Method & Syn. & RSTPReid \\
      \midrule
      HAP\cite{yuan2023hap} & \xmark & 49.4 \\
      $\rm{HAP^{\dag}}$\cite{yuan2023hap} & \xmark & 54.4 \\
      \midrule
      $\rm{StableRep_{CC}}$\cite{tian2023stablerep} & \checkmark & 43.0 \\
      $\rm{StableRep_{Red}}$\cite{tian2023stablerep} & \checkmark & 42.8 \\
      SynCLR\cite{tian2023synclr} & \checkmark & 21.8 \\
      SynthCLIP\cite{hammoud2024synthclip} & \checkmark & 21.7 \\
      \midrule
      $\rm{GenPoCCL_{COCO}}$ & \checkmark & \textbf{46.2} \\
      $\rm{GenPoCCL_{LU}}$ & \checkmark & 45.2 \\
      \bottomrule
    \end{tabular}
  \end{subtable}
  \hfill
  \begin{subtable}[t]{0.50\textwidth}
    \centering
    \subcaption{\textbf{Attribute recognition.} We evaluate on PA-100K\cite{liu2017pa100k} and PETA\cite{deng2014peta} datasets. mA($\%$) is reported.}
    \label{subtab:attribute}
    \begin{tabular}{lccc}
      \toprule
      Method & Syn. & PA-100K & PETA \\
      \midrule
      HAP\cite{yuan2023hap} & \xmark & 86.5 & 88.4 \\
      $\rm{HAP^{\dag}}$\cite{yuan2023hap} & \xmark & 81.8 & 87.9 \\
      \midrule
      $\rm{StableRep_{CC}}$\cite{tian2023stablerep} & \checkmark & 78.4 & \textbf{85.2} \\
      $\rm{StableRep_{Red}}$\cite{tian2023stablerep} & \checkmark & 77.4 & 84.5 \\
      SynCLR\cite{tian2023synclr} & \checkmark & 63.4 & 72.7 \\
      SynthCLIP\cite{hammoud2024synthclip} & \checkmark & 60.6 & 72.5 \\
      \midrule
      $\rm{GenPoCCL_{COCO}}$ & \checkmark & \textbf{79.0} & 85.0 \\
      $\rm{GenPoCCL_{LU}}$ & \checkmark & 77.8 & 83.6 \\
      \bottomrule
    \end{tabular}
  \end{subtable} 
\end{table}

\subsection{Main results} \label{subsec:main_result}
  In this section, we benchmark human-centric perception tasks to contrast the efficacy of models pre-trained with conventional methods using generated images, against those trained with our GenPoCCL.
  Notably, $\rm{GenPoCCL_{COCO}}$ requires only $\sim$0.6M generated samples, less than 1\% of the data volume used by StableRep (\cref{subtab:data_scale}). 
  Even with this smaller dataset, GenPoCCL succeeds in capturing structural features of the human body, outperforming StableRep in subsequent human-centric perception tasks.
  Also, as shown in \cref{tab:results}, increasing the scale of the dataset has the potential to further improve performance.
  While some tasks do not show improved performance, the reasons are discussed in detail in \cref{sec:limitation}.
  Please note that although the performance of HAP trained only with real images surpasses GenPoCCL, our goal is to utilize generated data for pre-training.
  Hence, the following section details a performance comparison between GenPoCCL trained on the GenCOCO dataset and StableRep trained on the GenCC12M dataset, focusing on evaluations across various tasks.

  \subsubsection{2D pose estimation.}
    The goal of this task is to localize human pose keypoints displayed in the image.
    \cref{subtab:pose} shows that GenPoCCL outperforms StableRep by +0.9\% on MPII\cite{andriluka2014mpii}, +0.1\% on MSCOCO\cite{lin2014microsoft}, respectively.
    Also, compared to an alternative approach which directly utilizes input conditions and corresponding generated images as direct paired training data, GenPoCCL outperforms +2.8\% on MSCOCO dataset.
    This result indicates that the consistency between input conditions and corresponding generated images is insufficient for direct supervision, yet adequate for indirect supervision.
  \subsubsection{Person ReID.}
    The purpose of this task is to retrieve a person of interest across multiple non-overlapping cameras. 
    \cref{subtab:reid} shows that GenPoCCL outperforms StableRep by +5.5\% on Market-1501\cite{zheng2015market} dataset.
    Interestingly, the gradient explodes for the StableRep pre-trained model in very early training iterations, so we clamp the training loss to stabilize training.
  \subsubsection{Text-to-image person ReID.}
    The goal of this task is to search for pedestrian images of a specific identity via textual descriptions.
    \cref{subtab:t2i_reid} shows that GenPoCCL outperforms StableRep by +3.2\% on RSTPReid\cite{zhu2021rstpreid} dataset.
  \subsubsection{Pedestrian attribute recognition.}
    The aim of this task is to assign multiple attributes to one pedestrian image.
    \cref{subtab:attribute} shows that GenPoCCL outperforms StableRep by +0.6\% on PA100K\cite{liu2017pa100k} dataset, and achieves comparable performance on PETA\cite{deng2014peta} dataset.
    We argue that since human body pose understanding is less important than appearance understanding in this task, GenPoCCL does not show remarkable performance.

\begin{table}[t]
  \caption{Ablation analysis. \textit{PoCCL}, and \textit{[POSE]} stand for pose-consistent multi-positive contrastive learning and [POSE] token, respectively.}
  \label{tab:ablation}
  \centering
  \begin{tabular}{lcc|cccc}
    \toprule
    method     & PoCCL & [POSE] & MPII          & Market-1501    & RSTPReid       & PA-100K        \\ \midrule
    baseline-HAP &       &            & 88.7          & 81.6          & 41.4          & 78.5          \\
    GenPoCCL-0 & $\checkmark$     &            & 89.0          & \textbf{83.2} & 39.4          & 78.4          \\
    GenPoCCL   & $\checkmark$     & $\checkmark$          & \textbf{89.2} & 83.0          & \textbf{46.2} & \textbf{79.0} \\
    \bottomrule
  \end{tabular}
\end{table}

\subsection{Ablation study} \label{subsec:ablation}
  Ablation studies for the proposed components are detailed in \cref{tab:ablation}.
  The baseline is denoted as Baseline-HAP, which employed HAP's default settings on the GenCOCO dataset.

  To demonstrate the efficacy of the proposed pose-consistent multi-positive contrastive learning, we trained a model by solely adding it to the baseline-HAP, which we refer to as GenPoCCL-0.
  In other words, GenPoCCL-0 employs a [CLS] token to learn both HAP's structure-invariant alignment and our proposed pose-consistent multi-positive contrastive loss.
  Compared with baseline-HAP and GenPoCCL-0, our proposed GenPoCCL results in +0.3\% on MPII, and +1.6\% on Market-1501.
  Unfortunately, the performance on RSTPReid decreased by 2.0\%.
  These results show that letting the [CLS] token both learn discriminative human features and to seek feature alignment across different image sharing the same pose is suboptimal, as discussed in \cref{subsec:posetk}.
  We will show in the following paragraph incorporating the [POSE] token is highly effective for addressing this problem.
    
  Compared with GenPoCCL-0 and 
  GenPoCCL, which incorporating [POSE] token, results in +6.8\% on RSTPReid, and +0.6\% on PA-100K.
  On the other hand, the performance on Market-1501 decreased by 0.2\%.
  In text-to-image ReID (RSTPReid), focusing on more detailed human features is necessary.
  Our proposed [POSE] token allows for  the extraction of both discriminative human features and human-pose related features, which is critical for this task.
  
  \begin{table}[t]
    \caption{Performance comparison of our proposed GenPoCCL with different data augmentation settings. 
             We additionally adopt several types of augmentations (\eg, random rotation, random blur, \etc) for stronger augmentation.
             We show that strong augmentation is beneficial to capture important representation during pre-training.
             }
    \label{tab:res_aug}
    \centering
    \begin{tabular}{l|ccccc}
      \toprule
      Augmentation    & MPII         & Market-1501    & RSTPReid       & PA-100K        \\ \midrule
      match \cite{he2022mae,yuan2023hap} & 87.7 & 75.8          & 42.0 & 76.5 \\
      stronger & \textbf{89.2} & \textbf{83.0}          & \textbf{46.2} & \textbf{79.0} \\
      \bottomrule
    \end{tabular}
  \end{table}

  As mentioned in \cref{subsec:settings}, we adopt stronger data augmentation than the common settings in prior research\cite{he2022mae,yuan2023hap}.
  As shown in \cref{tab:res_aug}, our proposed stronger data augmentation results in +7.2\% on Market-1501, and +4.2 \% on RSTPReid.
  We consider this is because stronger data augmentation bridges the domain gap between real and generated images.
  
  Overall, by integrating pose-consistent multi-positive contrastive learning and the addition of the [POSE] token, our GenPoCCL outperforms the baseline-HAP.
  Consequently, our proposed GenPoCCL outperformed StableRep in various human-centric perception tasks even with a smaller dataset, demonstrating significant benefits of the design of our GenPoCCL.

\begin{figure}[tbp]
  \begin{minipage}[b]{0.49\textwidth}
    \centering
    \includegraphics[width=\textwidth]{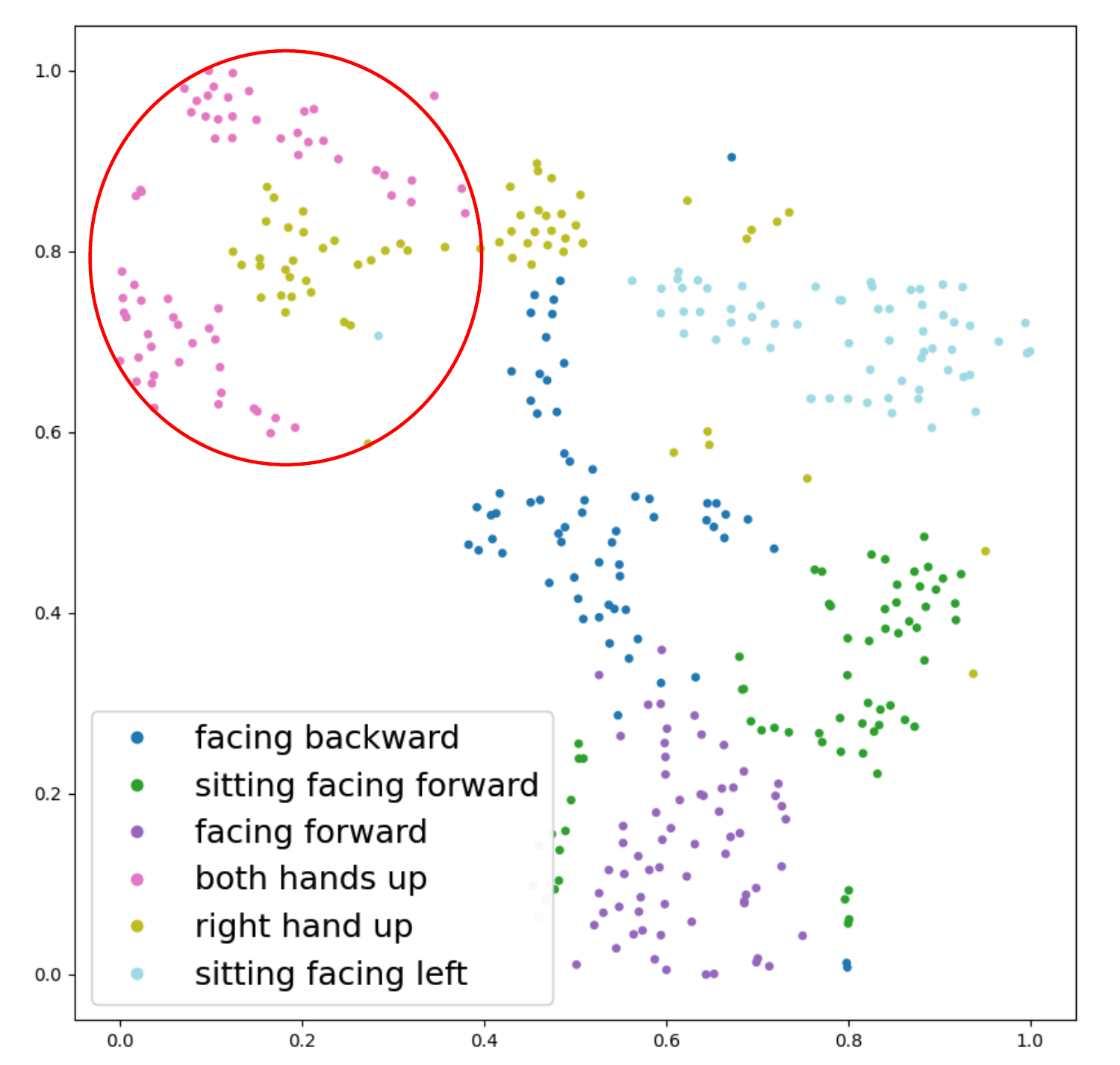}
    \subcaption{StableRep trained on GenCC12M dataset.}
  \end{minipage}
  \begin{minipage}[b]{0.49\textwidth}
    \centering
    \includegraphics[width=\textwidth]{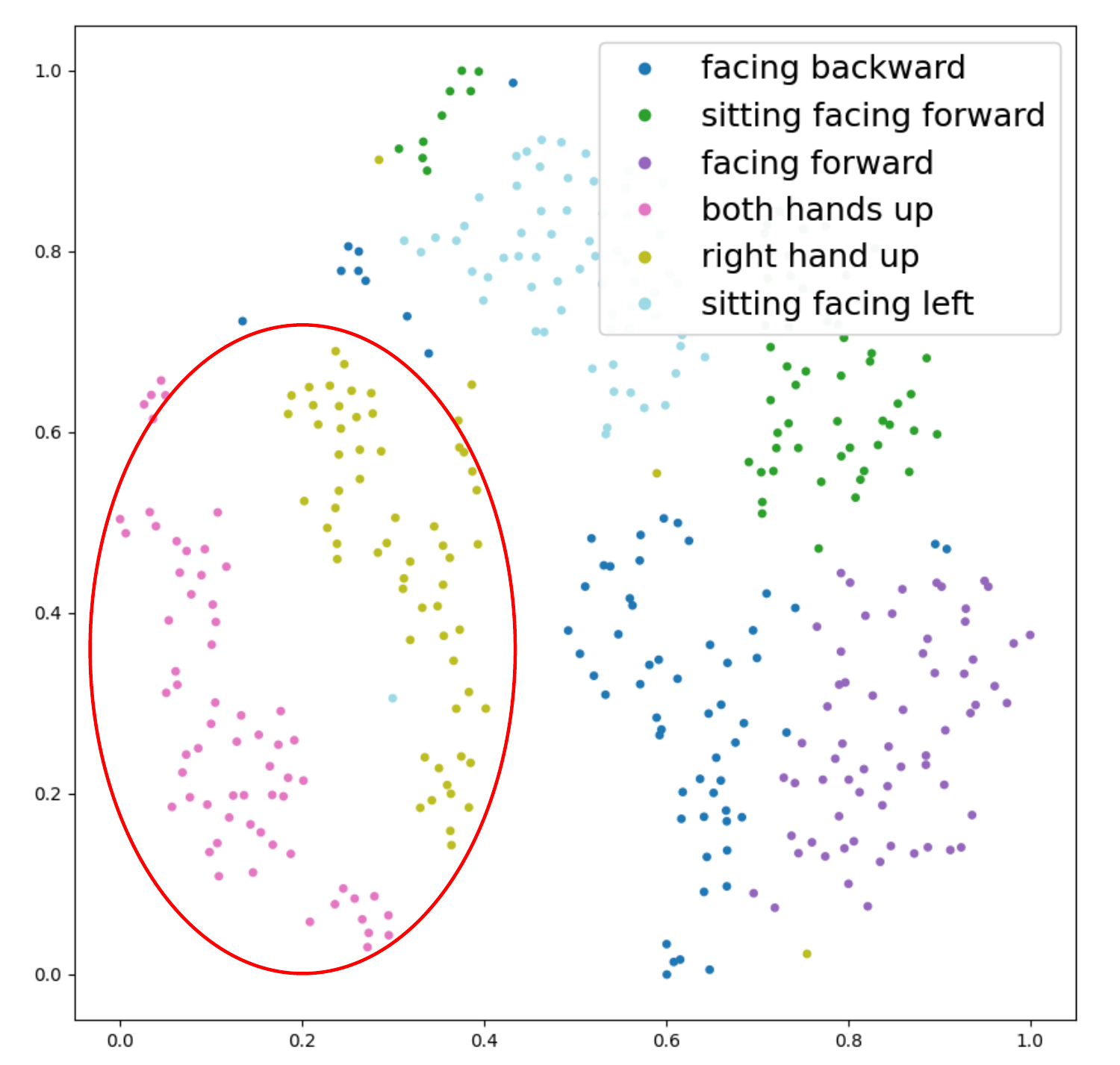}
    \subcaption{GenPoCCL trained on GenCOCO dataset.}
  \end{minipage}
  \caption{Comparison of t-SNE visualization between StableRep and GenPoCCL. As we can see, GenPoCCL is able to cleanly separate the features of different poses. It implies that GenPoCCL better captures human pose representation.}
  \label{fig:qualitative}    
\end{figure}

\subsection{Qualitative results} \label{sec:qualitative}
  We visualize intermediate features of the pre-trained feature extractor by utilizing t-SNE\cite{van2008visualizing}.
  Initially, for each of the six human pose pseudo-labels from the LUPerson dataset, we generate 64 images with consistent poses and varied appearances. 
  Subsequently, we extract and visualize their features using pre-trained models.
  As shown in \cref{fig:qualitative}, compared to the StableRep pre-trained model, our GenPoCCL pre-trained model is able to map images with similar poses to proximate spaces.
  As indicated by the red circles, the StableRep pre-trained model struggles to discriminate between poses with one hand raised and those with both hands raised, whereas our model succeeds in distinguishing between them.
  It means our GenPoCCL pre-trained model succeeded in learning the structural features of the human body.
  Consequently, our GenPoCCL pre-trained model easily adapts to various human-centric perception tasks, resulting in better performance.

\subsection{Limitations} \label{sec:limitation}
  While our method is proven to be effective, we recognize its limitations.
  Firstly, the quality of the generated images can be suboptimal, particularly in the case of human faces.
  Additionally, we do not specifically evaluate the quality of pose consistency between the generated image and the input condition.
  We consider these to be one of the reasons for the performance degradation compared to methods using real data.
  Furthermore, inconsistencies between generated images and provided captions or conditions may compromise the generated data's quality and utility.
  Additionally, our GenLUPerson dataset is created using a rule-based caption generation scheme, which is considered to be less than optimal.
  We consider that employing large language models (LLMs) for caption generation, akin to the SynCLR\cite{tian2023synclr} or SynthCLIP\cite{hammoud2024synthclip} approach, might enhance the quality and diversity of the synthetic captions.

\section{Conclusion}
  This paper presents the first investigation into pre-training solely with generated images for human-centric perception tasks and introduces a novel method named GenPoCCL.
  By generating pose-consistent, appearance-varying images and employing pose-consistent multi-positive contrastive learning to align their features,
  we experimentally demonstrated the ability to effectively capture the structural features of the human body, even with synthetic images.
  Even with merely less than 1\% of the data volume used by StableRep, GenPoCCL outperforms StableRep across various human-centric perception tasks.

\title{Supplementary Materials for Multi Positive Contrastive Learning with Pose-Consistent Generated Images}
\titlerunning{GenPoCCL}
\authorrunning{S.Inayoshi et al.}
\author{\empty}
\institute{\empty}
\maketitle

\renewcommand{\thesection}{\Alph{section}}
\renewcommand{\thefigure}{\Alph{figure}}
\renewcommand{\thetable}{\Alph{table}}
\setcounter{section}{0}
\setcounter{figure}{0}
\setcounter{table}{0}
\section{Details of multi-positive contrastive learning}
  The Pytorch-like pseudocode of the batched pose-consistent multi-positive contrastive learning algorithm is described in \cref{algo:MPCL}.
  We first assign consistent labels to images generated from the same human pose input.
  We then utilize cross-entropy loss during training to encourage the model to align the features corresponding to identical poses and distinctly separate the features from different ones.
  
  \begin{algorithm}[htbp]
    \SetAlgoLined
      \PyComment{latent vector z1: [n*m, l]} \\
      \PyComment{latent vector z2: [n*m, l]} \\
      \PyComment{n poses, m images per pose} \\
      \PyCode{def MPCLoss(z1, z2)} \\
      \Indp   
        \PyComment{compute ground-truth distribution p} \\
        \PyCode{label = range(n)} \\
        \PyComment{label images from same pose consistently by repeating m times} \\
        \PyCode{label = label[:, None].repeat(1, m).flatten()} \label{algo:MPCL:labelconsistent} \\
        \PyCode{p = (label.view(-1, 1) == label.view(1, -1))} \\
        \PyCode{p.fill\_diagonal(0)}  \PyComment{self masking} \\
        \PyCode{p /= p.sum(1)} \\
        \PyCode{} \\
        \PyComment{compute contrastive distribution q} \\
        \PyCode{logits = einsum(z1, z2) / tau} \\
        \PyCode{logits.fill\_diagonal(-1e9)}  \PyComment{self masking} \\
        \PyCode{q = softmax(logits, dim=1)} \\
        \PyCode{} \\
        \PyCode{H(p, q).backward()} \\
      \Indm 
      \PyCode{} \\
      \PyCode{def H(p, q):}  \PyComment{cross-entropy} \\
      \Indp
        \PyCode{return - (p * log(q)).sum(1).mean()} \\
      \Indm
    \caption{PyTorch-style pseudocode for multi-positive contrastive loss}
    \label{algo:MPCL}
  \end{algorithm}

\section{Details of strong data augmentation}
As introduced in the main paper, we adopt stronger data augmentation settings than existing methods\cite{he2022mae,yuan2023hap} for bridging domain gaps between real images and generated images.
We list detailed settings of data augmentation in \cref{tab:detail_aug}.

\begin{table}[htbp]
  \centering
  \begin{threeparttable}
    \caption{Details of data augmentation used in pre-training. 
    We use specific functions from Albumentations\cite{info11020125} package mentioned in \textit{augmentation} column.
    In addition, their parameters are listed in a \textit{parameters} column.
    }
    \label{tab:detail_aug}
    \centering
    \begin{tabularx}{\textwidth}{l|YY|YY}
      \toprule
      augmentation    & match\cite{he2022mae,yuan2023hap} & stronger & parameters & probability \\ \midrule
      ColorJitter & \xmark & \checkmark & \begin{tabular}{l}
        brightness=0.2\\contrast=0.2\\saturation=0.2\\hue=0.2 \end{tabular} & 0.8 \\ \midrule
      GaussianBlur & \xmark & \checkmark & \begin{tabular}{l} 
        blur\_limit=(3,7)\\sigma\_limit=0 \end{tabular} & 0.8 \\ \midrule
      ToGray & \xmark & \checkmark & - & 0.2 \\ \midrule
      Solarize & \xmark & \checkmark & threshold=128 & 0.2 \\ \midrule
      HorizontalFlip & \checkmark & \checkmark & - & 0.5 \\ \midrule
      ShiftScaleRotate & \xmark & \checkmark & \begin{tabular}{l}
        shift\_limit=0.0\\scale\_limit=0.0\\rotate\_limit=45\\interpolation=2\tnote{$\spadesuit$}\\border\_mode=0\tnote{$\blacklozenge$}\\value=pix\_mean\tnote{$\clubsuit$}
      \end{tabular} & 1.0 \\ \midrule
      RandomResizedCrop & \checkmark & \checkmark & \begin{tabular}{l}
        height=256\\width=192\\scale=(0.8,1.0)\\ratio=(3/8, 2/3)\\interpolation=2
      \end{tabular} & 1.0 \\ \bottomrule
    \end{tabularx}
    \begin{tablenotes}
      \footnotesize
      \item [$\spadesuit$] cv2.INTER\_CUBIC=2.
      \item [$\blacklozenge$] cv2.BORDER\_CONSTANT=0.
      \item [$\clubsuit$] The mean pixel value used for input normalization. \\
      In our settings, pix\_mean=(123.675, 116.28, 103.53).
    \end{tablenotes}
  \end{threeparttable}
\end{table}

\section{Hyperparameters of human-centric perception tasks}
\subsection{2D human pose estimation}
  We evaluate the proposed GenPoCCL for 2D human pose estimation on MPII\cite{andriluka2014mpii} and MSCOCO\cite{lin2014microsoft} datasets.
  Leveraging the MMPose codebase\cite{mmpose2020}, we incorporate ViTPose\cite{xu2022vitpose} following HAP\cite{yuan2023hap}.
  The input resolution is fixed at 256$\times$192 pixels.
  Data augmentations are random horizontal flipping, half body transformations, and random scaling and rotation, which are same settings as HAP.
  We use mean square error(MSE) as the loss function to minimize the difference between predicted and ground-truth heatmaps.
  Please refer to \cref{tab:2dpose} for the hyper-parameter specifications.

  \begin{table}
    \caption{Hyper-parameters of 2D human pose estimation.}
    \label{tab:2dpose}
    \centering
    \begin{tabularx}{\textwidth}{lYYYYYYY}
      \toprule
      dataset & batch size & epochs & learning rate & optimizer & weight decay & layer decay & drop path \\ \midrule
      MPII & 512 & 210 & 2.0e-4 & Adam & - & - & 0.30\\
      MSCOCO & 512 & 210 & 5.0e-4 & AdamW & 0.1 & 0.80 & 0.30\\ 
      \bottomrule
    \end{tabularx}
  \end{table}

\subsection{Person ReID}
  We evaluate the proposed GenPoCCL for person ReID on Market-1501\cite{zheng2015market} dataset.
  We leverage the codebase of HAP, which originates from MALE\cite{lu2022improving}.
  The input resolution is fixed at 256$\times$128 pixels.
  Data augmentations are resizing, random flipping, padding, and random cropping, which are same settings as HAP.
  We use a cross-entropy loss and a triplet loss with equal weights of 0.5.
  Please refer to \cref{tab:reid} for the hyper-parameter specifications.

  \begin{table}
    \caption{Hyper-parameters of person ReID.}
    \label{tab:reid}
    \centering
    \begin{tabularx}{\textwidth}{lYYYYYYY}
      \toprule
      dataset & batch size & epochs & learning rate & optimizer & weight decay & layer decay & drop path \\ \midrule
      Market-1501 & 64 & 100 & 8.0e-3 & AdamW & 0.05 & 0.40 & 0.10\\
      \bottomrule
    \end{tabularx}
  \end{table}

\subsection{Text-to-image person ReID}
  We evaluate the proposed GenPoCCL for text-to-image person ReID on RSTPReid\cite{zhu2021rstpreid} dataset.
  We leverage the codebase of HAP, which originates from LGUR\cite{shao2022learning}.
  The input resolution is fixed at 384$\times$128 pixels.
  Data augmentations are resizing and random horizontal flipping, which are same settings as HAP.
  BERT\cite{devlin2018bert} is utilized to extract text embeddings, which are then fed into bidirectional LSTM (Bi-LSTM)\cite{hochreiter1997long}.
  Feature dimensions of image and text embeddings are both fixed at 384.
  The BERT is frozen while Bi-LSTM and ViT are trained with a cross-entropy loss and a ranking loss.
  Please refer to \cref{tab:t2ireid} for the hyper-parameter specifications.

  \begin{table}
    \caption{Hyper-parameters of text-to-image person ReID.}
    \label{tab:t2ireid}
    \centering
    \begin{tabularx}{\textwidth}{lYYYY}
      \toprule
      dataset & batch size & epochs & learning rate & optimizer \\ \midrule
      RSTPReid & 64 & 60 & 1.0e-3 & Adam \\
      \bottomrule
    \end{tabularx}
  \end{table}

\subsection{Pedestrian attribute recognition}
  We evaluate the proposed GenPoCCL for pedestrian attribute recognition on PA-100K\cite{liu2017pa100k} and PETA\cite{deng2014peta} dataset.
  We leverage the codebase of HAP, which originates from Rethinking of PAR\cite{jia2021rethinking}.
  The input resolution is fixed at 256$\times$192 pixels.
  Data augmentations are resizing and random horizontal flipping, which are same settings as HAP.
  The binary cross-entropy loss is utilized for multi-class classification training.
  Please refer to \cref{tab:pedestrian} for the hyper-parameter specifications.

  \begin{table}
    \caption{Hyper-parameters of pedestrian attribute recognition.}
    \label{tab:pedestrian}
    \centering
    \begin{tabularx}{\textwidth}{lYYYYY}
      \toprule
      dataset & batch size & epochs & learning rate & optimizer & weight decay \\ \midrule
      PA-100K & 64 & 55 & 1.7e-4 & AdamW & 5.0e-4 \\
      PETA & 64 & 50 & 1.9e-4 & AdamW & 5.0e-4 \\
      \bottomrule
    \end{tabularx}
  \end{table}

\section{Caption generation method}
  As introduced in the main paper, we adopt rule-based method to generate captions for GenLUPerson dataset.
  Firstly, we prepare two types of formats of captions as shown in \cref{tab:format}.
  We choose one of them randomly based on the probability written in \cref{tab:format}.
  Then, we randomly sample expressions for each variable contained in the format.
  The candidates of each variable is shown in \cref{tab:variables}.
  Through this approach, we can obtain captions such as "An Asian girl wearing sky blue one-piece" to be used as inputs for image generation.

  \begin{table}[ht]
    \caption{Two types of rule-based formats used for generating captions.
    Each bracket represents a variable where the choices are detailed in \cref{tab:variables}.}
    \label{tab:format}
    \centering
    \begin{tabularx}{\textwidth}{cY}
      \toprule
      format & probability \\ \midrule
      (race) (identity) wearing (color1) (upper cloth) and (color2) (lower cloth) & 0.7 \\
      (race) (identity) wearing (color1) (clothing) & 0.3 \\
      \bottomrule
    \end{tabularx}
  \end{table}

  \begin{table}[ht!]
    \caption{Candidates of each variable. We try our best to ensure inclusivity across all races to enrich generational diversity for our work.}
    \label{tab:variables}
    \centering
    \begin{tabularx}{\textwidth}{cY}
      \toprule
      variables & choices \\ \midrule
      race & An Asian, An Australian, An African, An European, A North American, A South American \\ \midrule
      identity & man, woman, boy, girl \\ \midrule
      color1, color2 & 
      \scriptsize
      aurora, baby pink, begonia, bougainvillea, camellia, coral pink, dawn pink, flamingo, flesh pink, mallow, nail pink, old rose, pastel pink, peach pink, pink, rose pink, salmon pink, shell pink, shrimp pink, sunrise pink, 
      alizarin red, azalea, burgundy, carmine, cherry red, crimson, garnet, lacquer red, magenta, poppy red, raspberry, red, rose red, ruby, scarlet, shrimp red, signal red, strawberry, turquoise red, wine red,
      apricot, brick red, brunt sienna, carrot, cinnamon, coral red, nasturtium, orange, persimmon orange, pumpkin, tangerine, tiger lily, topaz, vermilion,
      bister, bronze, brown, buff, burnt umber, camel, chestnut brown, chocolate, coffee brown, copper, cork, hazel brown, henna, khaki, mahogany brown, maroon, raw-sienna, sand beige, sepia, tan, terracotta, umber,
      bamboo, beige, blond, buttercup yellow, canary, chartreuse yellow, flesh, fragrant olive, gold, jasmine, jaune brillant, lemon yellow, lime green, maize, marigold, mimosa, mustard, Naples yellow, primrose yellow, saffron yellow, straw, sunflower, yellow ocher, yellow,
      apple green, aquamarine, avocado, billiard green, bottle green, celadon, chartreuse green, cobalt green, emerald, ever green, forest green, grass green, green, iron blue, ivy green, jade, jasper green, laurel green, leaf green, malachite green, marine blue, moss green, olive, parrot green, peppermint green, sage green, spruce, teal green, turquoise blue, turquoise green, viridian,
      aqua, baby blue, blue, candy blue, cerulean, cobalt blue, cyan, delft blue, duck blue, fog blue, forget me not, frosty blue, gentian blue, hyacinth, hydrangea blue, indigo, ink blue, lapislazuli, majolica blue, midnight blue, moonlight blue, navy blue, Nile blue, oriental blue, peacock blue, powder blue, Prussian blue, royal blue, saxe blue, sky blue, smalt, ultramarine, wedgewood blue,
      amaranth purple, bellflower, claret, crocus, eggplant, fuchsia, grape, heliotrope, iris, lavender, lilac, mauve, mulberry, orchid, pansy, pearl gray, peony, purple, raisin, royal purple, taupe, Victoria violet, violet, wistaria,
      black, charcoal gray, cloud gray, dusky gray, ebony, eggshell, gray, gunmetal gray, ivory black, ivory, lamp black, milky white, pearl white, slate gray, snow white, taupe, white \\ \midrule
      upper cloth & t-shirt, polo shirt, dress shirt, sweater, hoodie, tank top, crop top, blouse, cardigan, jacket, coat, vest, blazer, bolero, poncho, shawl, jersey, sweatshirt, pullover, henley shirt, rugby shirt, turtleneck, peplum top, halter top, tube top, camisole, corset, bustier \\ \midrule
      lower cloth & jeans, trousers, shorts, skirt, leggings, joggers, sweatpants, chinos, cargo pants, capris, culottes, harem pants, palazzo pants, pencil skirt, mini skirt, maxi skirt, A-line skirt, pleated skirt, wrap skirt, sarong, bermuda shorts, hot pants, board shorts, cycling shorts, bell-bottoms, slacks, tights, kilt \\ \midrule
      clothing & dress, overalls, jumpsuit, romper, playsuit, one-piece, dungarees, unitard, catsuit, salopettes, boiler suit, coveralls, flight suit, kimono, sari, cheongsam, hanbok, kaftan, toga, tunic\\
      \bottomrule
    \end{tabularx}
  \end{table}

  \subsection{Discussions}
    In our study, we utilized captions specifying race to generate diverse images which reflect a variety of racial features, as they are employed in the dataset for training generative models.
    While we use terms referring to specific regions or ethnicities, it is strictly for scientific analysis and advancement in image generation technology, without any intent to promote stereotypes or biases.
    The terms are carefully chosen to respect and accurately represent the diversity of each region and ethnicity.
    These expressions are essential within the research context and are indispensable for achieving our objectives.
    We commit to adhering to ethical standards in research and respecting the dignity and diversity of all individuals.

%
%
\bibliographystyle{splncs04}
\bibliography{main}
\end{document}